CT-CLIP: a multimodal recognition method for apple leaf diseases in complex orchard environments

Lemin Liu[a,b], Fangchao Hu[a,b], Honghua Jiang[a,b,*] Yaru Chen[c], Limin Liu[d], Yongliang Qiao[e,*]

[a] *College of Information Science and Engineering, Shandong Agricultural University, Taian 271018, China*

[b] *Shandong Provincial Smart Agriculture Characteristic Laboratory, Taian 271018, China*

[c] *Centre for Vision Speech and Signal Processing (CVSSP), University of Surrey, Guildford GU2 7XH, United Kingdom*

[d] *School of Mechanical and Electronic Engineering, Shandong Agricultural Engineering College, Jinan 250100, China*

[e] *Australian Institute for Machine Learning, University of Adelaide, Adelaide SA 5000, Australia*

**Abstract:** In complex orchard environments, the phenotypic heterogeneity of different apple leaf diseases, characterized by significant variation among lesions, poses a challenge to traditional multi-scale feature fusion methods. These methods only integrate multi-layer features extracted by convolutional neural networks (CNNs) and fail to adequately account for the relationships between local and global features. Therefore, this study proposes a multi-branch recognition framework named **CNN-T**ransformer-**CLIP** (CT-CLIP). The framework synergistically employs a CNN to extract local lesion detail features and a Vision Transformer to capture global structural relationships. An Adaptive Feature Fusion Module (AFFM) then dynamically fuses these features, achieving optimal coupling of local and global information and effectively addressing the diversity in lesion morphology and distribution. Additionally, to mitigate interference from complex backgrounds and significantly enhance recognition accuracy under few-shot conditions, this study proposes a multimodal image-text learning approach. By leveraging pre-trained CLIP weights, it achieves deep alignment between visual features and disease semantic descriptions. Experimental results show that CT-CLIP achieves accuracies of 97.38% and 96.12% on a publicly available apple disease and a self-built dataset, outperforming several baseline methods. The proposed CT-CLIP demonstrates strong capabilities in recognizing agricultural diseases, significantly enhances identification accuracy under complex environmental conditions, provides an innovative and practical solution for automated disease recognition in agricultural applications.

**Keywords:** Apple leaf disease; Disease recognition; CLIP; Multimodal learning;

## 1　Introduction

Apples rank among the most widely cultivated fruits in temperate climates due to their delicious taste and high nutritional value, placing third in global fruit production behind tomatoes and bananas. In 2022, worldwide apple output reached nearly 96 million metric tons, with approximately half of this total produced in China. (Arnold and Gramza-Michalowska, 2024; Fariñas-Mera et al., 2025; Markó et al., 2017; Schusterova et al., 2025; Yan et al., 2024). Apple leaves are vulnerable to various diseases, including rust and brown spot (Singh et al., 2022; Wöhner and Emeriewen, 2019), leads to significant yield reductions and economic losses (Liu et al., 2017).Traditional expert-based diagnosis is labor-intensive, prone to subjective bias, and thus lacks accuracy. Accurate disease identification is critical for curbing spread (Bankar et al., 2014). With the fast development of deep learning and computer vision, plant leaf disease recognition methods based on CNNs and ViTs have been widely applied (Cándido-Mireles et al., 2023; Das et al., 2025; Fariñas-Mera et al., 2025; Khotimah et al., 2023; Kussul et al., 2017; Shammi et al., 2023; Singh et al., 2021). Ji et al (Ji et al., 2020) integrated multiple CNNs models to enhance feature extraction and achieved 98.57% classification accuracy on the PlantVillage dataset. Zhang et al (Zhang et al., 2021) developed a module featuring multiple activation functions and preprocessing techniques to attain 97.41% accuracy on a self-constructed dataset. Yu and Son (Yu and Son, 2020) incorporated an attention mechanism to deep learning framework for leaf disease recognition specifically favorable for detecting spotting leaf lesions. Wang et al (Wang et al., 2022) developed a backbone network based on Swin Transformer, which improved identification of leaf diseases in real-world scenarios. Liu et al. (Liu et al., 2022) presented a two-stage PSPNet–UNet framework to diagnose the severity of Alternaria leaf disease, yielding an accuracy of 96.41%. Tian et al. (Tian et al., 2022) designed a V-space-based multi-scale feature fusion SSD named VMF-SSD for apple leaf disease detection, which constructed localization branch on the ground of V-space and a multi-scale feature fusion for improvement. A study was proposed to fine-tune the CNN model using a new type of multispectral (RGB-NIR) time series created specifically for this purpose to classify apple black spot disease symptoms(Bleasdale and Whyatt, 2025). The above studies demonstrated good performance on self-constructed or relatively simple datasets (e.g., images captured with fixed backgrounds), but often show reduced accuracy in complex outdoor environments.

In real apple orchards, the phenotypic heterogeneity of lesions, their high visual similarity, multi-stage evolution process, and blurred boundaries significantly impair the performance of identification models (Xiao et al., 2020). For example, brown spot and leaf spot diseases may be difficult to distinguish during certain growth stages. Additionally, lesion features are influenced by environmental factors such as lighting conditions, humidity, and leaf position (Khan et al., 2022). They can cause the same disease to display different visual characteristics under varying conditions. These factors make disease recognition based on single laboratory background images challenging to maintain stable accuracy in complex natural environments.

The aforementioned high degree of visual uncertainty indicates that relying solely on the image modality has become a fundamental bottleneck for enhancing model robustness. To break through this limitation, multimodal learning, which integrates multiple information sources, demonstrates significant potential. As shown in Fig. 1, visual-language models (VLMs), such as CLIP (Radford et al., 2021), connect visual and textual modalities through joint pre-training, thereby generating more robust and semantically informative feature representations. Although agricultural research currently relies primarily on visual data for disease classification(Yu et al., 2023), integrating multimodal information holds promise for enhancing interpretability and accuracy in real-world diagnostics. Therefore, this study proposes a novel multimodal feature aggregation framework, CT-CLIP. The method combines the pre-trained CNNs and ViTs of CLIP, integrating the local and global features. It also enhances feature learning through text, achieving the fusion of information across different modalities. To enhance efficient feature integration between the CNN and Transformer branches, AFFM was allocated to achieve seamless integration of local and global features. Additionally, inspired by GLIP (Li et al., 2022), Feature Enhancer Module that utilizes Bi-MultiHead Attention (BiMA) to strengthen interactions between image and text modalities. The main contributions of this study include three main components:

(1) We propose a dual-branch CT-CLIP architecture with an adaptive feature fusion module, which integrates CNNs and ViTs to jointly capture local disease details and global structural patterns. This design improves adaptability to diverse lesion morphologies and complex spatial distributions.

(2) We introduce contrastive learning to align visual features with textual disease descriptions in a unified semantic space. This alignment enhances robustness to background interference and improves the recognition of rare diseases.

(3) The CT-CLIP model achieves recognition accuracies of 97.38% and 96.12% on public and self-built datasets, effectively addressing challenges of diverse symptoms and complex environments, providing solid technical support for intelligent orchard management.

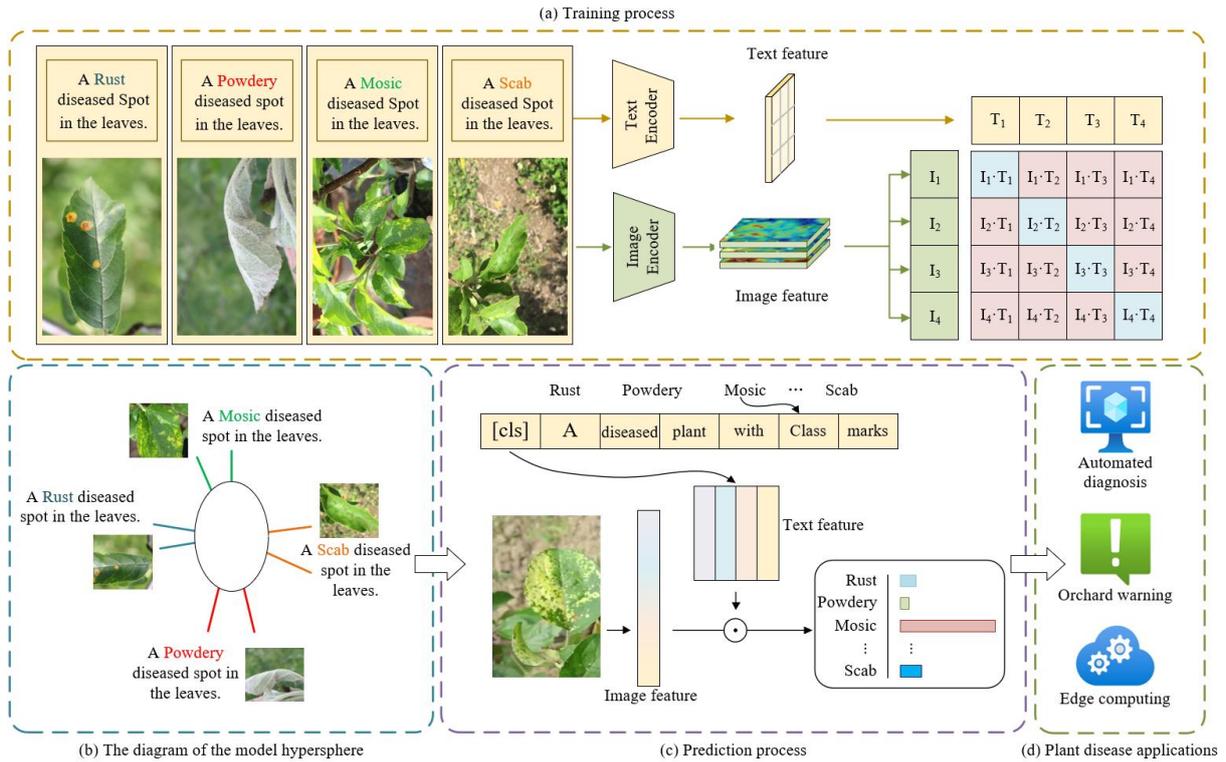

**Fig. 1.** (a) Training Process: Images and text descriptions of plant diseases (Rust, Powdery, Mosaic, and Scab) are encoded separately and combined to form feature matrices for training. (b) Model hypersphere: Within this model hypersphere space, paired text disease instances are located in close proximity. (c) Prediction Process: The model uses encoded features to predict disease classes. (d) Application scenarios: The system will support automated diagnosis, orchard early warning, and edge computing for real-time disease monitoring in the future.

## 2   Materials and methods

### 2.1   Multi-source complex real-orchard dataset

To evaluate the model's generalization capability in real-world complex environments, this study employs a multi-source apple leaf disease image repository integrating both public datasets and self-collected data. This composite dataset combines controlled laboratory settings with complex field scenarios, designed to simulate and address the core challenges in practical agricultural applications.

The dataset utilized (as shown in Table 1) integrates images of five disease categories (26,377 images in total) constructed by Northwest A&F University and the AppleLeaf9 dataset (14,066 images, 94% of which were captured in field environments) released by Southwest University. Furthermore, to enhance data diversity and evaluate model performance under authentic regional conditions, an independent collection was conducted at the Science and Technology Innovation Park of Shandong Agricultural University (117.16°E,

36.16°N). The images were captured between 8:00 AM and 4:00 PM daily using an iPhone 13 Pro Max, with a resolution of 3024 × 4032 pixels. After a rigorous screening process, a final dataset containing 1,413 RGB images covering five common leaf diseases was established.

Table 1 Distribution of different disease categories in the dataset.

| Class | Public Datasets | Self-built Datasets |
|---|---|---|
| Alternaria | 5760 | 256 |
| Brown Spot | 6066 | 260 |
| Grey | 5149 | - |
| Mosaic | 5246 | - |
| Rust | 8447 | 307 |
| Powdery | 1184 | 311 |
| Scab | 5410 | 279 |

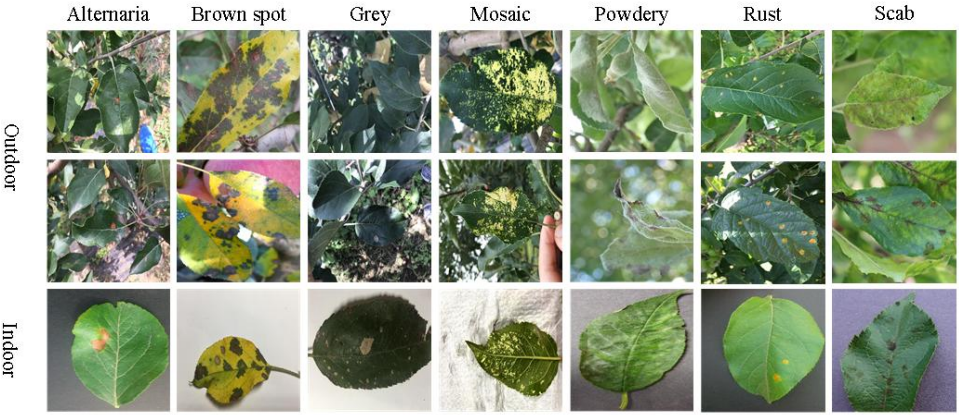

(a) The public dataset employed in this study

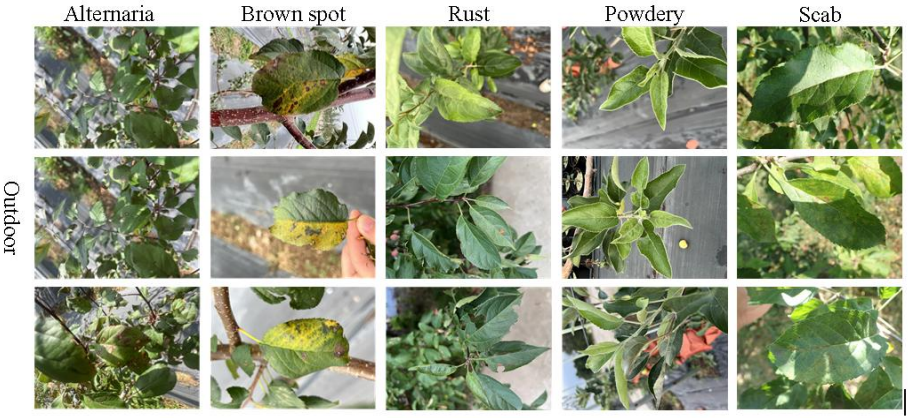

(b) dataset collected from apple orchard

Fig. 2. Example images of apple disease leaves

This multi-source dataset exhibits significant inter-class similarity and intra-class variability, including inconsistent image quality and complex background variations. More importantly, the disease representations range from localized small lesions (e.g., Rust) to large-scale diffuse lesions (e.g., Powdery Mildew), posing substantial challenges to the model's ability to simultaneously capture local details and global contextual information. Representative samples of the dataset are shown in Fig. 2.

## 2.2 Methodology

### 2.2.1 CT-CLIP based plant leaf disease classification

To address the challenge of simultaneously capturing fine-grained local symptoms and their global contextual associations, this study proposes the CT-CLIP architecture, as shown in Fig. 3. The input images are encoded by CLIP-CNN and CLIP-ViT encoders to extract local and global features of the images. To enhance recognition of rare diseases, the text input is encoded by a text encoder BERT to extract text features. Then these features are fused in the AFFM, and the parameters are adjusted by the adapters. Next the image and text features are processed in the FEB to ensure the effective interaction of image and text features, processing in the fully connected layer, the model outputs the corresponding category predictions, completing the classification task.

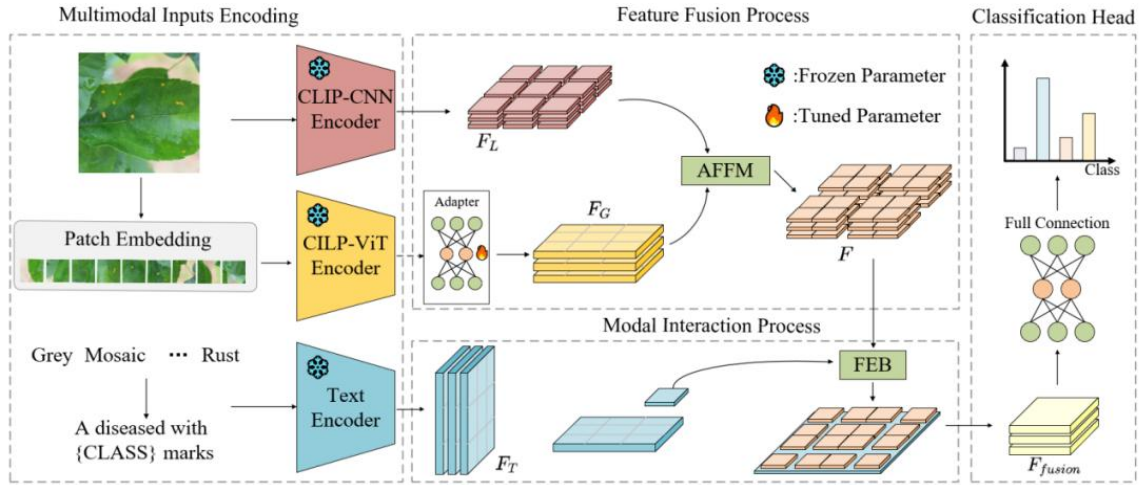

**Fig. 3.** The CT-CLIP architecture primarily consists of three components: Multimodal Input Encoding, Feature Fusion Process, and Modal Interaction Process.

Given an input batch $I \in \mathbb{R}^{B \times C \times H \times W}$ and perform feature extraction separately using CLIP-pretrained ResNet50 and ViT. The CNN branch outputs feature $F_L \in \mathbb{R}^{B \times C \times H \times W}$, while the ViT branch outputs features $F_G \in \mathbb{R}^{B \times N \times C}$ where $N$ represents the number of patches. To align the two branches, we reshape $F_G$ to dimensions $B, C, H, W$ and concatenate it with $F_L$ as the input to the AFFM, producing the fused feature $F \in \mathbb{R}^{B \times N \times C}$. On the text side, each class name $C_i$ is inserted into a predefined prompt template $T$ to generate prompt texts. These prompts are encoded using a CLIP-pretrained BERT model, yielding $T \in \mathbb{R}^{B \times L \times C}$. The [CLS] token is extracted as the global text feature $F_T \in \mathbb{R}^{B \times 1 \times C}$ Subsequently, the image feature $F$ and text feature $F_T$ are fed into the FEB for cross-modal feature fusion and enhancement. The process follows the equations (1–3):

$$T = BERT(Tokenizer([T; C_i])) \tag{1}$$

$$F_L = ClipCNN(I) \ ; \ F_G = ClipVit(I) \tag{2}$$

$$F = AFFM(F_G, F_L) \tag{3}$$

where $ClipCNN(\cdot)$ denotes the ResNet50 encoder pretrained by $ClipVit(\cdot)$ represents the ViT encoder pretrained by CLIP, and $AFFM(\cdot,\cdot)$ refers to the proposed feature fusion method.

### 2.2.2 Multimodal inputs encoding

To enable the model to rapidly adapt to disease phenotype heterogeneity with minimal parameter tuning, for a given an input image-text pair $I - T$, feature extraction is performed on the image using the CLIP pre-trained ResNet50 and ViT, while the text encoder uses BERT to extract features from the text $T$. This study utilize the pretrained CLIP weights, the ViT branch freezes most of the parameters in the CLIP model and only trains a lightweight Adapter Layer within the Transformer layers (Houlsby et al., 2019). The adapter module adjusts the distribution of pretrained features with a small number of trainable parameters, enabling the model to converge quickly. To deal with overfitting and enhance model robustness, CT-CLIP employs residual connections to dynamically fuse fine-tuned knowledge with the original knowledge of the CLIP backbone. Fig. 4 illustrates the architecture of the adapter and its integration within the Transformer.

The learnable adapter $A_I(\cdot)$ is inserted between the sub-layer output projection and the residual connection. Its architecture employs two linear transformations and a nonlinear activation to project the feature dimension $d$ to a smaller dimension $r$, then restore it back to the original dimension $d$. The total number of parameters is $2dr + r + d$. A constant $\alpha$ is used as the "residual ratio" to retain the original knowledge and strike a balance between performance and parameter efficiency. As shown in Equations (4) and (5):

$$A_I(F_G) = ReLU(F_G^T \mathbf{W}_1^v)\mathbf{W}_2^v \tag{4}$$

$$F_G^\star = \alpha A_I(F_G)^T + (1 - \alpha)F_G \tag{5}$$

After obtaining the updated image feature $F_G^\star$ the usage procedure is as equations (1) computing the class probability vector $p_i = \{p_i\}_{i=1}^K$. The predicted class of the image is the one with the highest probability, denoted $\hat{i}$, and expressed as $\hat{i} = \arg\max_i p_i$. During training, the weight of $A_I(\cdot)$ is optimized by reducing the cross-entropy loss in Equation (6):

$$\mathcal{L}(\theta) = -\frac{1}{N}\sum_{n=1}^{N}\sum_{i=1}^{K} y_i^{(n)} \log \hat{y}_i^{(n)} \tag{6}$$

where $N$ is the total number of training samples. When $i$ equals the true class label $\hat{i}$, $y_i = 1$; otherwise, $y_i = 1$. $\hat{y}_i = p_i$ denotes the predicted probability of class $i$. $\theta = \{\mathbf{W}_1^v, \mathbf{W}_2^v\}$ represents all the trainable parameters.

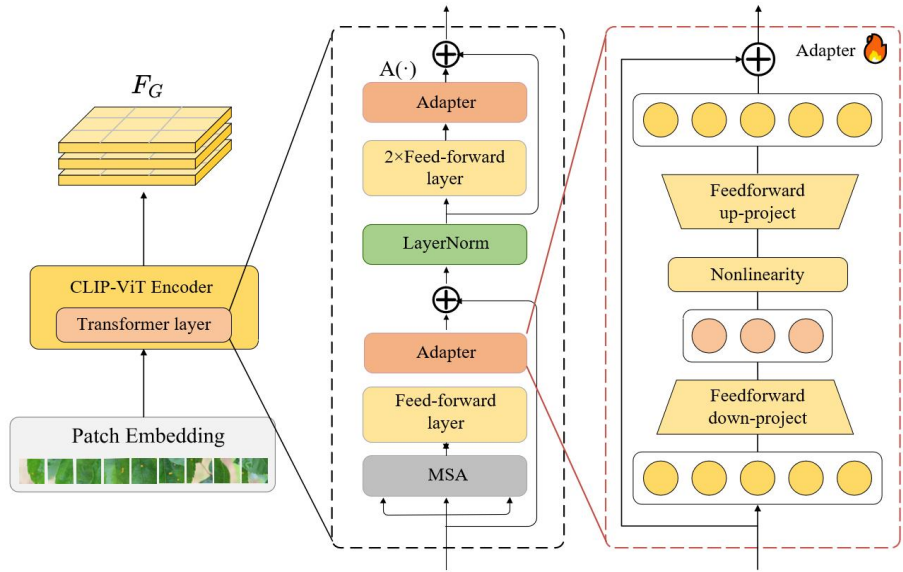

**Fig. 4.** The architecture of the adapter module and its integration with the Transformer

### 2.2.3 Feature fusion process

While the aforementioned CLIP-pretrained dual-branch visual encoder possesses extensive prior knowledge and cross-modal understanding capabilities, it struggles to accurately balance the influence of local and global features when processing complex apple leaf disease images. Thus, after obtaining features $F_L \in \mathbb{R}^{B \times C \times H \times W}$ and $F_G \in \mathbb{R}^{B \times N \times C}$ from two feature spaces, convolutional blocks are first used to further extract features and adjust dimensions to obtain $X$. This is then fed into the AFFM, which dynamically adjusts feature weights to enhance the representation of key features. Finally, the fused features undergo further convolutional processing and dimensional adjustments to obtain the final feature $F$. The AFFM module can adaptively learn and adjust the weights of local and global features, thereby enabling better feature fusion. Details of the AFFM module steps and parameters are provided in Table 2.

Specifically, the AFFM module consists of a Dynamic Attention Module (DAM) and a fusion component shown Fig. 5. The DAM first applies Adaptive Average Pooling to the input feature $X$ to extract global context information. Then, a fully connected layer and ReLU activation function are used to perform nonlinear mapping and feature transformation. After normalization via SoftMax, two sets of dynamic attention weights ($att^1$ and $att^2$) are generated based on the input features. These dynamic attention weights are applied via element-wise multiplication to the original features, enhancing the model's ability to focus on important features. Its processing allows the model to adaptively highlight salient regions or key features, thereby optimizing feature representation and improving the model's overall performance.

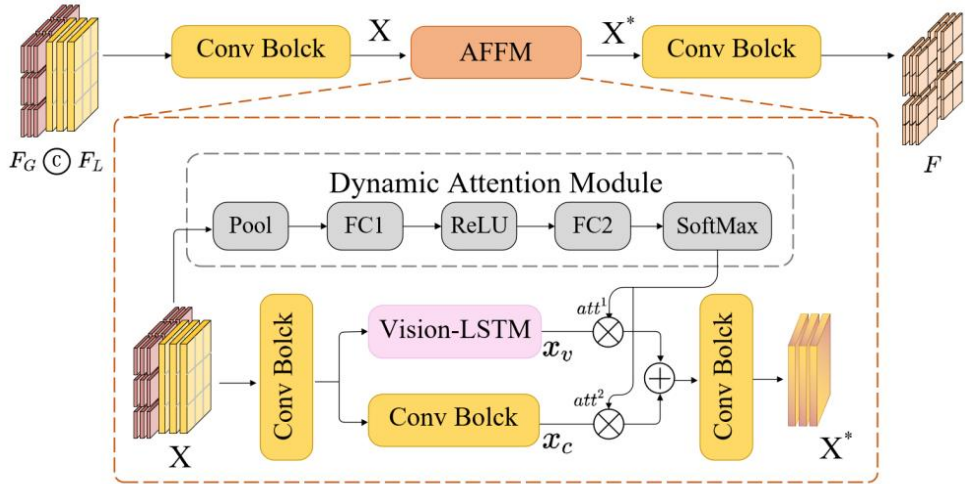

**Fig. 5.** Architecture of AFFM: It enhances the representational capacity of critical features by dynamically adjusting feature weights from different branches.

The fusion component introduces the Vision-Long Short Term Memory network (V-LSTM) (Alkin et al., 2024). V-LSTM is capable of constructing implicit sequences in image data, processing local features to capture long-range dependencies, thereby obtaining a portion of the global features. By incorporating V-LSTM, the module can more comprehensively model the dynamic relationships between features. Finally, the output features of the dynamic attention module are fused and further enhanced through convolution layers and the ReLU activation function, generating the final output features $X^*$, as shown in Equation (7).

$$X^* = f_{conv}(att_1 \times x_v + att_2 \times x_c) \tag{7}$$

where, $x_v$ represents the output of the V-LSTM branch, and $x_c$ represents the output of the CNN branch. $att\cdot$ represents the weights assigned to each branch. These features include multi-scale information, long-range dependency information, and the weighted results of dynamic attention.

**Table 2** Shows the operations corresponding to different processes in AFFM and their output dimensions. DAM refers to the processes within the Dynamic Attention Module, while Fusion represents the processes in fusion component.

| Step | Operate | Shape |
| --- | --- | --- |
| Input | $X$ | $(B, C, H, W)$ |
| DAM: Avg_Pooling | $Avgpool(x).view(a, b)$ | $(B, C)$ |
| DAM: MLP(FC1+ReLU+FC2) | $Linear() + ReLU() + Linear()$ | $(B, 2)$ |
| DAM: SoftMax | $att = SoftMax()$ | $(B, 2)$ |
| Fusion: Conv + ReLU | $Conv: 3 \times 3, p = 1, s = 1$ | $(B, C, H, W)$ |
| Fusion: $x_v$ | $Vision - LSTM\ Block()$ | $(B, C, H, W)$ |
| Fusion: $x_c$ | $Conv: 3 \times 3, p = 1, s = 1$ | $(B, C, H, W)$ |
| Fusion: Fusion | $x_v \cdot att^1 + x_c \cdot att^2$ | $(B, C, H, W)$ |
| Fusion: Conv + ReLU | $Conv: 3 \times 3, p = 1, s = 1$ | $(B, C, H, W)$ |
| Output | $X'$ | $(B, C, H, W)$ |

### 2.2.4 Model interaction process

In the Model interaction process, to improve the interaction between image and text features, this study introduces the FEB, which employs a Bi-MultiHead Attention mechanism to effectively model the interaction between modalities in Fig. 6. This facilitates the fusion of information across different modalities and strengthens the integration of image and text features. Specifically, let the visual features be represented as $v \in \mathbb{R}^{B \times N_v \times d_v}$ and the text features as $l \in \mathbb{R}^{B \times N_l \times d_l}$, where $B$ is the batch size, $N_v$ and $N_l$ are the sequence lengths of visual and text features, respectively, and $d_v$ and $d_l$ denote their corresponding feature dimensions. It applies linear transformations to the input visual and text features to obtain the Query (Q), Key (K), and Value (V) matrices. The projection process is defined as Equation (8):

$$Q_v = W_q^v v, \quad K_l = W_k^l l, \quad V_v = W_v^v v, \quad V_l = W_v^l l \tag{8}$$

where $W_q^v \in \mathbb{R}^{d_v \times d}$、$W_k^l \in \mathbb{R}^{d_l \times d}$、$W_v^v \in \mathbb{R}^{d_v \times d}$ 和 $W_v^l \in \mathbb{R}^{d_l \times d}$ are learnable parameter matrices, and $d$ is the shared embedding dimension. Next, the embedded features are split into multiple attention heads, with each head having a dimension of $d_h = d/H$, where $H$ represents the number of attention heads. The transformed feature shapes are $Q_v, K_l, V_v, V_l \in \mathbb{R}^{B \times H \times N \times d_h}$, where $N = N_v$ for visual features and $N = N_l$ for text features.

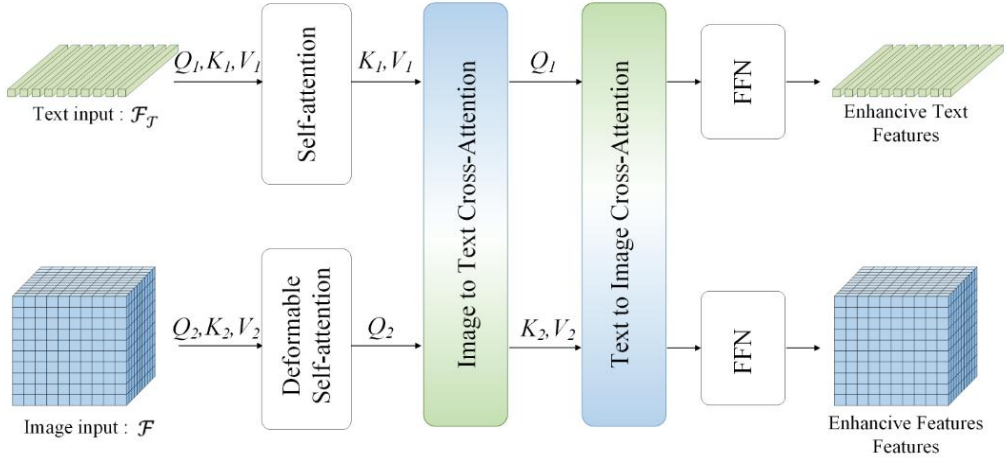

**Fig. 6.** FEB: Feature enhancement through BiMA mechanism

In the visual-to-language direction, the similarity between the visual Query and the linguistic Key is calculated, followed by a scaled dot-product attention, as described in Equation (9). Similarly, in the language-to-visual direction, computing the similarity between the linguistic Query and the visual Key, as shown in Equation (10):

$$A_v = \text{softmax}(\frac{Q_v K_l^T}{\sqrt{d_h}}) \quad (9)$$

$$A_l = \text{softmax}\left(\frac{Q_l K_v^T}{\sqrt{d_h}}\right) \quad (10)$$

where $A_v \in \mathbb{R}^{B \times H \times N_v \times N_l}$ 和 $A_l \in \mathbb{R}^{B \times H \times N_l \times N_v}$ represent the cross-modal attention weight matrices for the visual-to-language and language-to-visual interactions, respectively.

The study utilizes the attention weights to perform a weighted sum over the Value matrices, obtaining the final attention outputs: $O_v = A_v V_l$, $O_l = A_l V_v$, $O_v \in \mathbb{R}^{B \times H \times N_v \times d_h}, O_l \in \mathbb{R}^{B \times H \times N_l \times d_h}$. Next, concatenate the multi-head outputs and apply a final linear transformation to obtain the output features $\hat{v}$ and $\hat{l}$, as given in Equation (11):

$$\hat{v} = W_o^v \text{concat}(O_v), \hat{l} = W_o^l \text{concat}(O_l) \quad (11)$$

where $W_o^v, W_o^l \in \mathbb{R}^{d \times d_v}$ are projection matrices. The final output shapes are $\hat{v} \in \mathbb{R}^{B \times N_v \times d_v}$ and $\hat{l} \in \mathbb{R}^{B \times N_l \times d_l}$.

Then, $\bar{v}$ and $\bar{l}$ are obtained from $\hat{v}$ and $\hat{l}$ by global mean pooling, as shown in Equation (12). Finally, a weighted fusion is performed to generate $F_{\text{fusion}}$, as given in Equation (13). The fused features $F_{\text{fusion}}$ is then passed through a fully connected layer (FC) and classified into seven categories using Equation (14):

$$\bar{v} = Pooling(\hat{v}) \in \mathbb{R}^{B \times d_v}, \bar{l} = Pooling(\hat{l}) \in \mathbb{R}^{B \times d_l} \quad (12)$$

$$F_{\text{fusion}} = W_v \bar{v} + W_l \bar{l} \quad (13)$$

$$y = softmax(FC_2(ReLU(FC_1(F_{fusion})))) \quad (14)$$

where $W_v$ and $W_l$ are learnable weights, $FC_1 \in \mathbb{R}^{(d_v+d_l) \times d_h}$ is the hidden layer projection, and $FC_2 \in \mathbb{R}^{d_h \times 7}$ is the final apple leaves disease classification layer.

## 3 Experimental setup

### 3.1 Implementation details

The experiments were performed in an Ubuntu 18.04 environment using a GeForce GTX 4090 GPU, with PyTorch 1.13.0 serving as the deep learning framework. Randomly split the training and test datasets using an 8:2 ratio. The initial learning rate was set to 0.000035 and dynamically adjusted during training to accelerate model convergence. The batch size was set to 64, with a momentum of 0.9 and a weight decay coefficient of 1e-4. The total number of training iterations was 200. The optimizer used was Adam, replacing the traditional stochastic gradient descent (SGD). Batch Normalization was applied to standardize the input of each hidden layer. For visual encoders and text encoders, the visual encoder uses ViT-B/16 and ResNet50 based on CLIP pre-training, and the text encoder uses BERT. During training, the template "A diseased plant with {Class} marks" was used, where Class represents different category labels. The parameter settings for the other models involved in the experiments in Section 4.1 are shown in Table 3. To assess the model's accuracy and generalization capability in a few-shot setting, the dataset comprises seven disease categories, as shown in Fig. 2(a), with 300 samples per class. The data were split into training and testing sets at an 8:2 ratio for model training and evaluation. Subsequently, the model was tested on a self-built dataset to verify its generalization capability, as shown in Fig. 2(b).

**Table 3** The training parameters of involved networks, where "CE Loss" is Cross-Entropy Loss.

| Parameters | ResNet50 | ResNet152 | ViT-B/16 | Swin TransV2 |
| --- | --- | --- | --- | --- |
| Batch size | 32 | 32 | 32 | 32 |
| Loss function | CE Loss | CE Loss | CE Loss | LabelSmooth |
| Optimizer | SGD | SGD | AdamW | AdamW |
| Learning rate | 0.0005 | 0.0005 | 1e-4 | 0.0005 |

| Updater | Step | Step | Cosine Annealing | Cosine Annealing |
|---|---|---|---|---|

### 3.2 Evaluation metrics

This study evaluates the performance of CT-CLIP using four metrics: Precision (P), Recall (R), F1-Score, and Accuracy (Acc). For the multi-class classification task, these metrics are first calculated for each disease category individually, and then averaged to obtain the overall performance. Specifically, the total number of True Positives (TP), True Negatives (TN), False Negatives (FN), and False Positives (FP) is computed globally. Based on these values, the performance of the proposed model was evaluated using precision, recall, and F1-Score. These metrics are calculated as equations (15-18). Among them, the Acc metric evaluates the model's overall prediction performance, calculated as the proportion of accurately predicted samples to the total number of samples.

$$Accuracy = \frac{TP+TN}{TP+TN+FP+FN} \times 100\% \tag{15}$$

$$Precision = \frac{TP}{TP+FP} \times 100\% \tag{16}$$

$$Recall = \frac{TP}{TP+FN} \times 100\% \tag{17}$$

$$F1-score = \frac{2 \times (Precision \times Recall)}{Precision+Recall} \times 100\% \tag{18}$$

## 4 Results and discussion

### 4.1 Comparison of different methods

In the comparative experiments, this study utilizes CLIP with pre-trained weights from 400 million image-text pairs and fine-tunes the model parameters. To validate the effectiveness of the proposed model, several classic and state-of-the-art architectures in recent years (ResNet-152, EfficientNetV2-XL, PoolFormer-S, VAN) were selected for comparison (Guo et al., 2023; He et al., 2016; Tan and Le, 2019; Yu et al., 2022). The experimental results are shown in Table 4. Fig. 7 shows the heat map of model performance indicators. It can be seen that CT-CLIP outperforms other models in all metrics. Fig. 8 shows the F1-score contour plot of P and R. From this, it can be concluded that AlexNet, IResNet, and SDINet all exhibit high precision and low recall, accurately identifying typical symptoms but missing some global diseases. In contrast, the method used in this study achieves near-perfect balance between P and R. The accuracy and loss curves, smoothed curves, and earning rate decline trend during training are illustrated in Fig. 9. Accuracy, Loss, and learning rate stage analysis are shown in Fig. 10. We can see that the accuracy and loss have converged to a large extent by the epoch 50, and have almost converged after the epoch 100, thanks to the large number of pre-trained parameters in CLIP. The confusion matrix of the method used in this study and the current mainstream method-PoolFormer-S are shown in Fig. 11. The model's loss function and evaluation metrics stabilize after approximately 80 epochs, demonstrating that the prior knowledge provided by CLIP pre-training effectively enhances both training efficiency and generalization capability.

**Table 4** Comparison results of different models on public datasets.

| Methods | Acc (%) | P (%) | R (%) | F1 (%) |
| --- | --- | --- | --- | --- |
| AlexNet | 87.12 | 89.42 | 81.55 | 85.30 |
| Resnet152 | 92.10 | 92.10 | 92.10 | 92.11 |
| MEAN-SSD | 91.57 | 90.31 | 89.88 | 90.09 |
| LWCNN | 90.05 | 88.61 | 87.43 | 88.02 |
| IResNet | 87.20 | 86.24 | 84.15 | 85.18 |
| SDINet | 94.11 | 93.58 | 92.80 | 93.19 |
| EfficientNetV2-XL | 94.75 | 94.87 | 94.75 | 94.76 |
| PoolFormer-S | 96.10 | 91.93 | 91.90 | 91.91 |
| **CT-CLIP** | **97.38** | **97.38** | **97.41** | **97.38** |

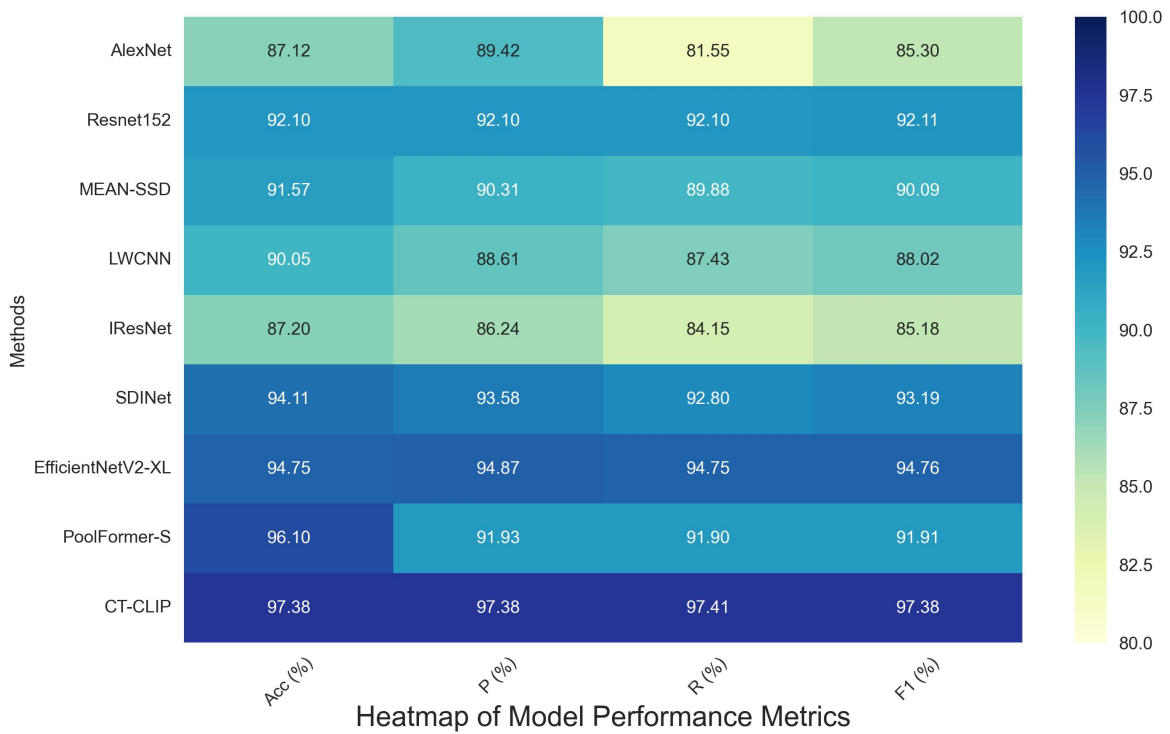

**Fig. 7.** Heat map of model performance metrics. Colors represent model performance on that metric.

CT-CLIP outperforms all eight other models across various performance metrics. It achieves an accuracy of 97.38%, a recall of 97.38%, a precision of 97.41%, and F1-Score of 97.38%, demonstrating significant improvements compared to PoolFormer-S, with accuracy increases of 1.28%. This performance improvement is primarily attributed to two key innovations. The network architecture adopts a multi-branch structure, effectively integrating local and global features, which significantly enhances disease recognition under complex and variable conditions. By incorporating textual modality information, the model maximizes the alignment between text and image feature spaces, providing semantic guidance that further improves classification accuracy. However, while other methods perform a four-class classification task, the proposed method in this study expands the task to seven classes by introducing three additional disease categories. This significantly increases data complexity and makes the classification task more challenging.

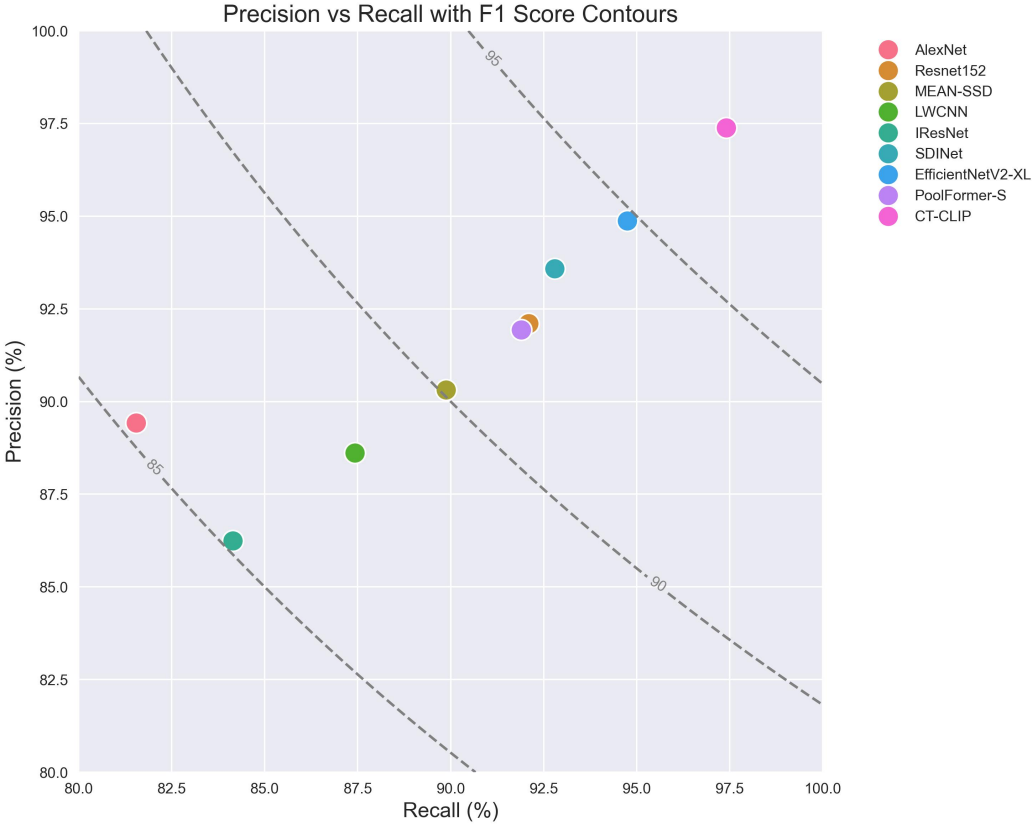

**Fig. 8.** F1 score contour plot of precision and recall rates. The upper left area represents a high false negative rate, while the lower right area represents a low precision rate.

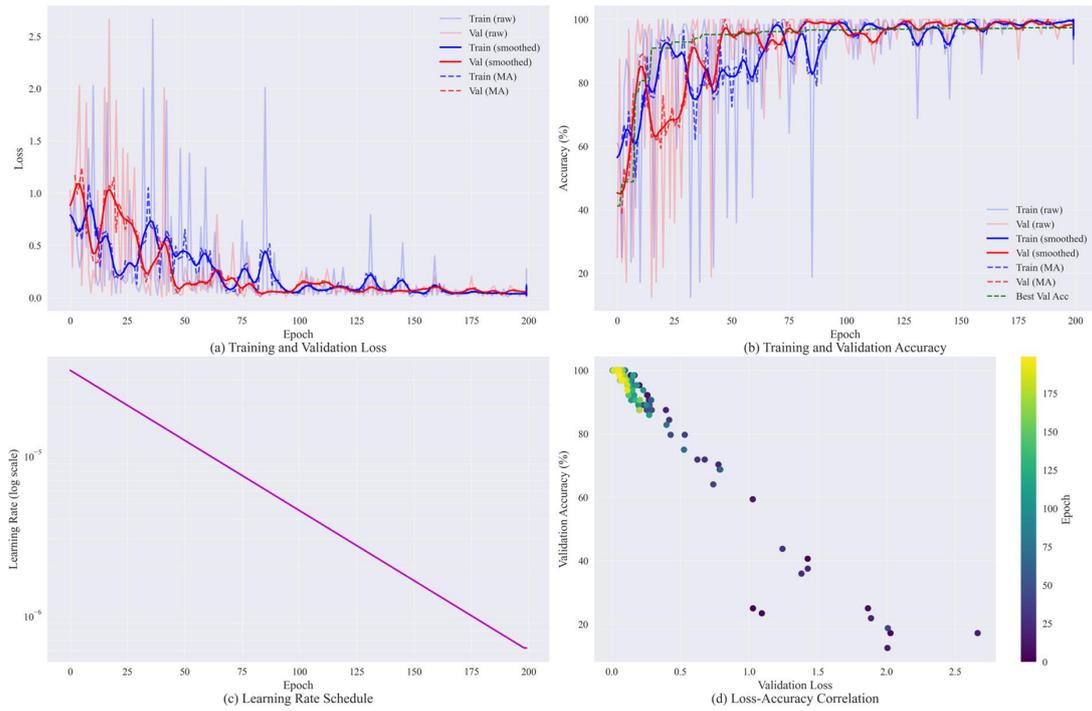

**Fig. 9.** Accuracy and Loss curves, learning rate decline trend, and Correlation analysis between loss and accuracy rate.

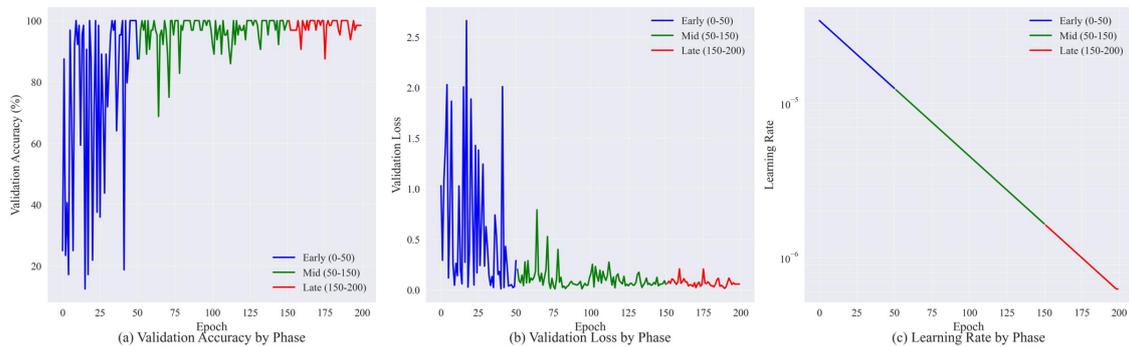

**Fig. 10.** Comparison of accuracy rates across three stages and verification of the effect of learning rate decay by stage. (a) and (b) show the accuracy rates and loss changes across different stages, while (c) shows the loss under different learning rates across different stages.

To further investigate the classification capability of CT-CLIP for images with similar disease symptoms, Fig. 12 shows the classification results for difficult samples. The results indicate that the method proposed in this study can classify correctly in indoor and outdoor environments, as shown in Fig. 12 (a), but there are a few cases with low-probability identification errors, as shown in Fig. 12 (b). This is because some disease morphologies are highly similar, and the environmental background is complex, leading to severe visual bias.

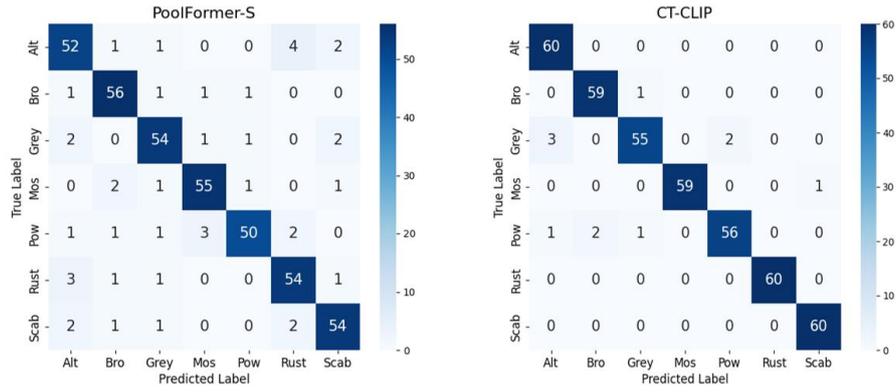

**Fig. 11.** Classification confusion matrix of CT-CLIP and PoolFormer-S on public datasets

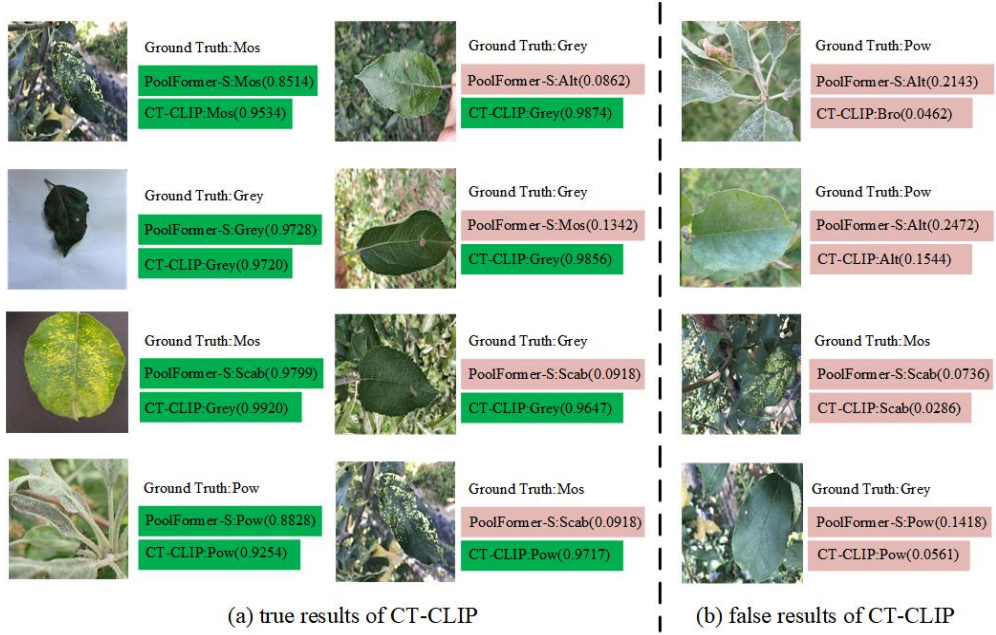

**Fig. 12.** Classification results for some disease samples in the public datasets. Red indicates classification error, green indicates correct classification, and the numbers indicate classification probability.

**4.2 Generalization experiments on self-built datasets**

To validate the generalization performance of our method, this study conducted generalization experiments using the self-built dataset introduced in Section 2.1.2 and compared it with current mainstream models (ResNet-50, ResNet-152, ViT-B/16, SwinTransV2-Base). Then. compared the results across four metrics: Accuracy (Acc), Precision (P), Recall (R), and F1-Score. The experimental results are shown in Fig. 13. the classification of difficult samples is shown in Fig. 14.

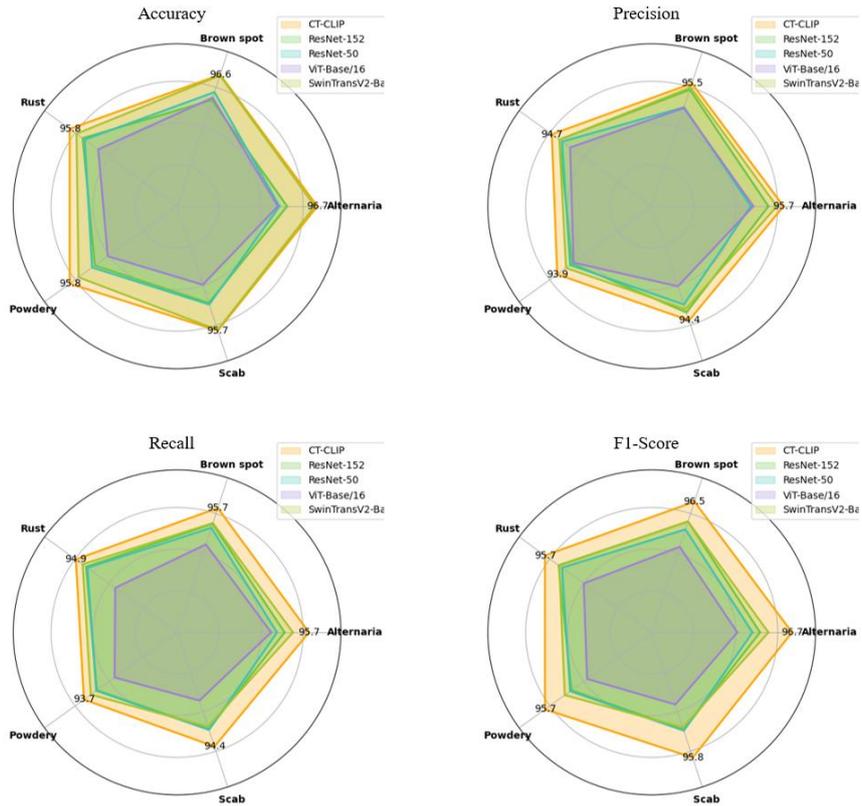

**Fig. 13.** Comparison of CT-CLIP results with multiple mainstream models on a self-built dataset, the circles of different colors representing different models.

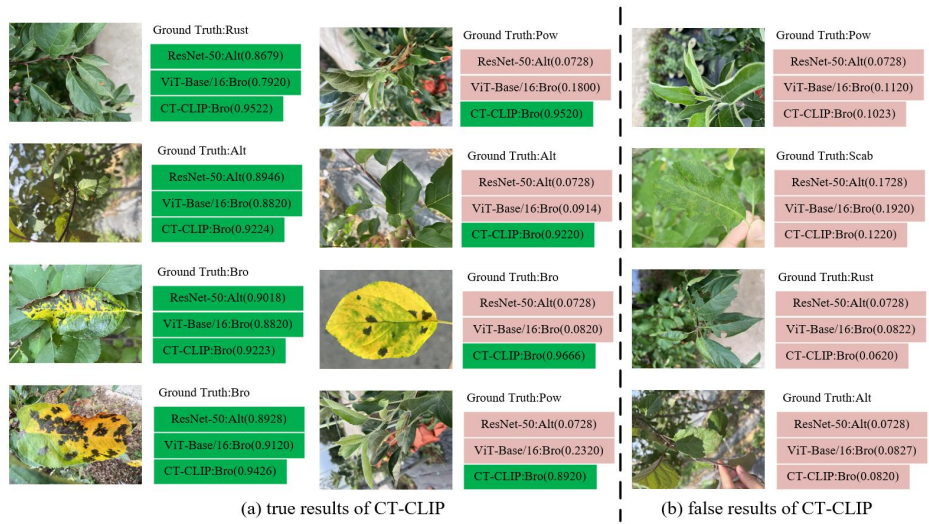

**Fig. 14.** Classification results for some disease samples in the self-built datasets. Red indicates classification error, green indicates correct classification.

CT-CLIP demonstrates outstanding performance in accuracy, precision, recall, and F1-Score. This superior performance stems from its effective integration of visual and textual information, which enhances feature representation through multimodal learning and improves discrimination of subtle inter-category differences. In contrast, while SwinTransV2-

Base excels in recall, its focus on identifying more positive samples results in a slight decrease in precision. It may perform well in capturing local features but lacks in understanding global features. The relatively poor performance of the ResNet series (50, 152) and ViT in this task may be related to their feature extraction methods not fully matching the feature distribution of the dataset, especially ResNet-152, which may suffer from overfitting due to its large number of layers, thereby affecting its generalization ability.

### 4.3 Ablation experiments

This section conducted three sets of ablation experiments to verify the effects of each module on CT-CLIP, the impact of the internal modules of DAM on feature fusion, and the different attention mechanisms used in the modal interaction process. Fig. 15 shows the comprehensive ablation results.

#### 4.3.1 Performance analysis of each modules

To evaluate the performance of each module, a systematic ablation study was conducted. Based on the baseline model, the Text, Adapter, AFFM, and FEB modules were progressively integrated into the CT-CLIP model to verify their contributions to performance. The experimental results are shown in Table 5.

**Table 5** Comparison of the impact of different modules on recognition results in two different datasets.

| Text | CNN | ViT | ViT (W/oAdapter) | FEB | AFFM | Accuracy (%) | |
|---|---|---|---|---|---|---|---|
| | | | | | | Public | Self-built |
| | √ | | | | | 93.57 | 93.21 |
| | | | √ | | | 92.85 | 92.15 |
| | | √ | | | | 93.81 | 93.86 |
| | √ | √ | | | | 94.76 | 93.96 |
| | √ | √ | | | √ | 95.71 | 95.64 |
| √ | √ | √ | | √ | | 96.43 | 96.08 |
| √ | √ | √ | | √ | √ | **97.38** | **96.12** |

**Note:** √ means that this module is in use.

From Table 5, it can be observed that the single-branch architecture ViT (W/o Adapter) performed the worst, with an public datasets and self-built datasets of 92.85% and 92.15%. After adding the Adapter module, the performance has improved by 0.96% and 1.71%, indicating that training the Adapter module enables the model to learn more information with minimal additional parameters. The dual-branch architecture, which incorporates both CNN and ViT, demonstrates a notable enhancement in model accuracy, achieving improvements of 1.19% and 0.95% respectively over single-branch architectures that employ CNN or ViT independently. Furthermore, adding the AFFM module to the dual-branch architecture, which integrates features from both CNN and ViT spatial domains, resulted in a public datasets and self-built datasets increase of 0.95% and 1.68%. On top of this, introducing the text modality and incorporating the FEB module for cross-modal feature fusion further boosted public datasets and self-built datasets by 1.67% and 2.12%, indicating integrated textual and visual modalities can compensate for the limitations of single-modal image features and significantly enhance model accuracy.

### 4.3.2 Performance evaluation of different submodules in the AFFM

To comprehensively evaluate the impact of the AFFM module on the overall model architecture, a series of detailed experiments were conducted. These experiments consist of two parts: one is to verify the effectiveness of DAM, and the other is to analyze the impact of V-LSTM and ViT on the AFFM model. The experimental results are shown in Table 6.

**Table 6** The performance of different branches in AFFM on feature fusion.

| CNN | V-LSTM | ViT | DAM | Accuracy (%) | |
| --- | --- | --- | --- | --- | --- |
| | | | | Public | Self-built |
| √ | | | | 95.89 | 95.39 |
| √ | √ | | | 96.54 | 95.61 |
| √ | | √ | | 96.11 | 94.68 |
| √ | √ | | √ | **97.38** | **96.12** |

**Note:** √ means that this module is in use.

After incorporating V-LSTM, the performance on public and self-built datasets improved by 0.65% and 0.22%, respectively. In contrast, replacing V-LSTM with a standard ViT yielded smaller improvements illustrated in Table 6. This demonstrates that ViT and V-LSTM may have similar capabilities in modeling long-range dependencies. Although the performance gap between V-LSTM and ViT is marginal, V-LSTM's gating mechanism offers advantages in simplifying model training. Furthermore, adding the DAM module further improved public datasets and self-built datasets by 0.84% and 0.51%, indicating that the DAM module effectively addresses the issue of uneven feature distribution across different spatial domains by providing dynamic weights

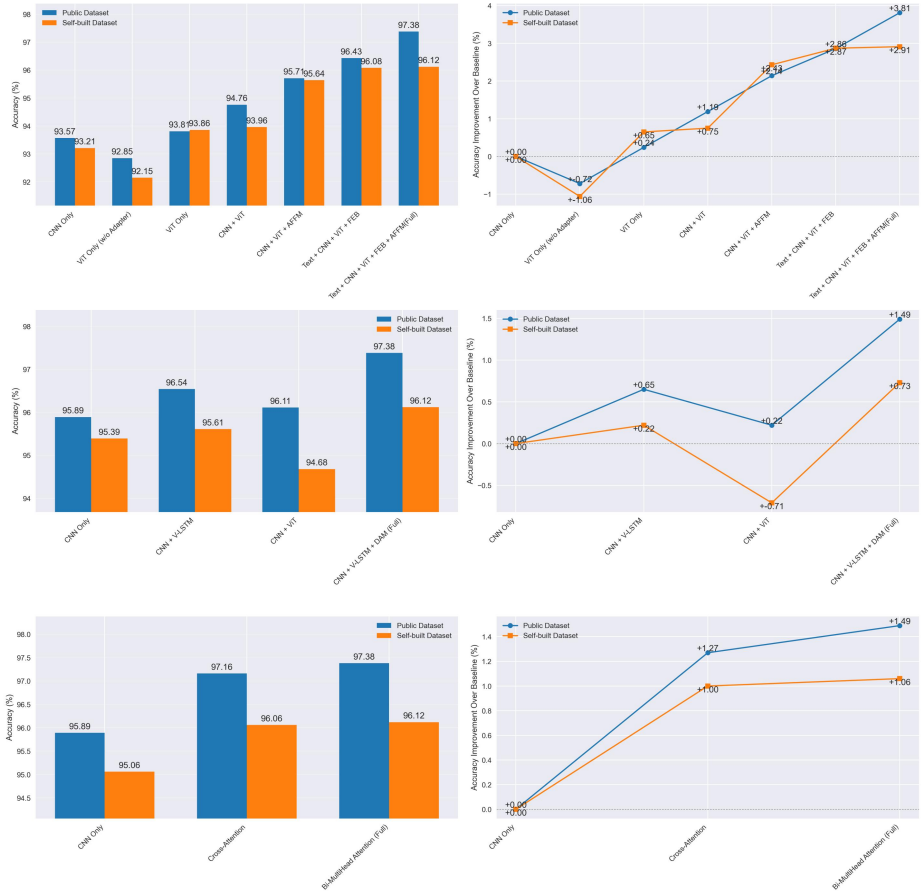

**Fig. 15** Attention heatmaps for three disease types.

### 4.3.3    Performance comparison of different attention mechanisms in FEB

To evaluate the performance of various attention mechanisms on FEB, we conducted a comparative study by substituting its attention layer with three alternatives: a CNN, cross-attention, and Bi-MultiHead Attention (BiMA). As shown in Table 7, the BiMA mechanism outperformed the other two approaches, achieving improvements of 0.32% and 1.49% on public datasets. These results suggest that BiMA's two-step attention process not only refines textual features but also facilitates more effective interaction and fusion between textual and visual features.

**Table 7** The performance of different attention mechanisms in FEB on classification accuracy.

| CNN | Cross-Attention | Bi-MultiHead Attention | Accuracy (%) | |
| --- | --- | --- | --- | --- |
| | | | Public | Self-built |
| √ | | | 95.89 | 95.06 |
| | √ | | 97.16 | 96.06 |
| | | √ | **97.38** | **96.12** |

**Note:** √ means that this module is in use.

### 4.4 Experiments visualization

To further demonstrate the performance of each module in CT-CLIP on performance improvement, this study conducted feature map visualization using sample data from the dataset. By examining important regions in the feature extraction layers, and gain deeper insights into the features and patterns learned by the model. In the section, this study uses Grad-CAM (Selvaraju et al., 2017) to visualize the model's focus regions during apple leaf disease identification. To provide a clearer comparison of model performance, selecting three distinct disease types for analysis, with the visualization results shown in Fig. 16.

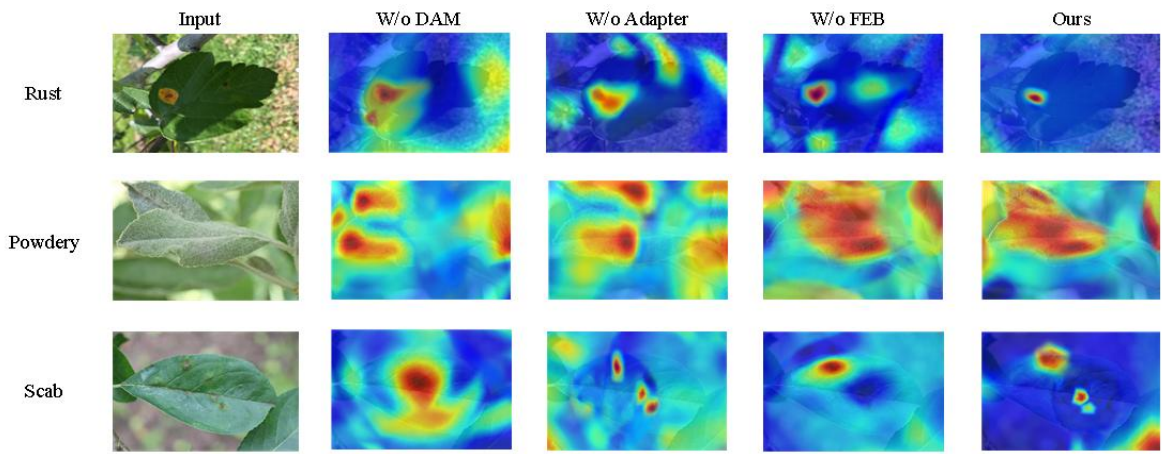

**Fig. 16** Attention heatmaps for three disease types.

The attention heatmap is generated based on the self-attention scores between the visual category token and all patch tokens in the visual encoder. Fig. 16 highlights fine-grained information emphasized by global features, where redder regions indicate stronger attention to specific local details. Our method consistently focuses on the diseased spots, whereas the attention distribution becomes more scattered after removing certain modules. It demonstrates that the CT-CLIP architecture effectively integrates features from different receptive fields, enabling better attention to disease-affected regions and ultimately enhancing model performance.

## 5 Conclusions and future works

To address the challenges of distinct symptom patterns, background interference, and limited sample sizes in complex orchard environments, this study proposes a multi-branch recognition framework (CT-CLIP) that aggregates multimodal features. By constructing a dual-branch visual encoder combining convolutional neural networks with Transformers, coupled with an adaptive feature fusion module, we effectively overcome the limitation of single models struggling to simultaneously capture both fine-grained local disease features and global contextual associations. Second, the introduction of an image-text contrastive learning mechanism maps visual features and disease semantic descriptions onto the same vector space. This approach not only enhances recognition stability for common diseases through prior knowledge but also significantly improves generalization capabilities for rare and emerging diseases.

Experimental results demonstrate that this model achieves an accuracy of 97.38% on a public apple disease dataset and 96.12% on a self-built dataset, surpassing mainstream baseline methods and exhibiting robust generalization capabilities. This provides a viable pathway for advancing plant disease recognition technology in smart agriculture from laboratory research to practical field applications. Future work will explore integrating hyperspectral and video information while incorporating lightweight architectures to enhance the model's adaptability for field deployment and industrial application value, further advancing precision agriculture.

**CRediT authorship contribution statement**

**Lemin Liu:** Writing - original draft, Methodology, Visualization. **Fangchao Hu:** Writing - review & editing. **Yongliang Qiao:** Methodology, Supervision, Writing - review & editing. **Honghua Jiang:** Investigation, Methodology, Supervision, Funding acquisition, Writing - review & editing. **Yaru Chen:** Validation, Data curation, Writing - review & editing. **Limin Liu:** Writing - review & editing.

**Declaration of Competing Interest**

The authors declare that they have no known competing financial interests or personal relationships that could have appeared to influence the work reported in the study.

**Acknowledgements**

This work was supported by the Shandong Provincial Natural Science Foundation (grant numbers ZR2023MC201); Key Research and Development Program of Shandong Province (grant numbers 2024TZXD043).